\documentclass[11pt]{article}
\usepackage{graphicx}
\usepackage{longtable,lscape}
\usepackage{amsthm}
\usepackage{amsfonts}
\usepackage{amsmath}
\usepackage{bbm}
\usepackage{float}
\usepackage{url}

\newcommand{\captionfonts}{\footnotesize}
\makeatletter  % Allow the use of @ in command names
\long\def\@makecaption#1#2{%
  \vskip\abovecaptionskip
  \sbox\@tempboxa{{\captionfonts #1: #2}}%
  \ifdim \wd\@tempboxa >\hsize
    {\captionfonts #1: #2\par}
  \else
    \hbox to\hsize{\hfil\box\@tempboxa\hfil}%
  \fi
  \vskip\belowcaptionskip}
\makeatother
\begin{document}
\title{Modeling Concept Combinations in a Quantum-theoretic Framework}
\author{Diederik Aerts and Sandro Sozzo \vspace{0.5 cm} \\ 
        \normalsize\itshape
        Center Leo Apostel for Interdisciplinary Studies \\
        \normalsize\itshape
        and, Department of Mathematics, Brussels Free University \\ 
        \normalsize\itshape
         Krijgskundestraat 33, 1160 Brussels, Belgium \\
        \normalsize
        E-Mails: \url{diraerts@vub.ac.be,ssozzo@vub.ac.be}
           \\
              }
\date{}
\maketitle
\begin{abstract}
\noindent
We present modeling for conceptual combinations which uses the mathematical formalism of quantum theory. Our model faithfully describes a large amount of experimental data collected by different scholars on concept conjunctions and disjunctions. Furthermore, our approach sheds a new light on long standing drawbacks connected with vagueness, or fuzzyness, of concepts, and
puts forward a completely novel possible solution to the `combination problem' in concept theory. Additionally, we introduce an explanation for the occurrence of quantum structures in the mechanisms and dynamics of concepts and, more generally, in cognitive and decision processes, according to which human thought is a well structured superposition of a `logical thought' and a `conceptual thought', and the latter usually prevails over the former, at variance with some widespread beliefs.
\end{abstract}

\section{Conceptual vagueness and the combination problem\label{intro}}
According to the `classical view', going back to Aristotle, a concept is considered practically as a `container of its instantiations'. This view was already criticized by Wittgenstein but definitely put at stake by Rosch's work on color, which showed that subjects rate concept membership of an instance as graded (or fuzzy, or vague). Hence mathematical notions to model such conceptual fuzziness were put forward. But, Osherson and Smith's examples of concept conjunctions revealed a fundamental difficulty of classical (fuzzy) set-theoretic approaches to model such conjunctions. These authors considered the concepts {\it Pet} and {\it Fish} and their conjunction {\it Pet-Fish}, and observed that, while an exemplar such as {\it Guppy} is a very typical example of {\it Pet-Fish}, it is neither a very typical example of {\it Pet} nor of {\it Fish}. Hence, the typicality of a specific exemplar with respect to the conjunction of concepts shows an unexpected behavior from the point of view of classical set and probability theory. That the `Pet-Fish problem' (also known as `Guppy effect') indeed revealed a fundamental conflict with classical structures, was confirmed in a crucial way by Hampton's studies \cite{h1988a,h1988b} which measured the deviation from classical set-theoretic membership weights of exemplars with respect to pairs of concepts and their conjunction or disjunction. Hampton showed that people estimate membership in such a way that the membership weight of an exemplar of a conjunction (disjunction) of concepts is higher (lower) than the membership weights of this exemplar for one or both of the constituent concepts. This phenomenon is referred to as `overextension' (`underextension'). Several experiments have since been performed and many elements have been taken into consideration with respect to this `combination problem' to provide a satisfactory mathematical model of concept combinations. Notwithstanding this, a model that represents the combination of two or more concepts from the models that represent the individual concepts still does not exist.

Meanwhile, it has been shown that quantum structures are systematically present in domains of the social sciences, e.g., in the modeling of cognitive and decision processes \cite{aertsgabora2005a,aertsgabora2005b,aerts2009,pb2009,k2010,bpft2011,bb2012,hk2013}. As such, we have developed a specific quantum-theoretic approach to model and represent concepts \cite{aertsgabora2005a,aertsgabora2005b,aerts2009,aertssozzo2011,aertsbroekaertgaborasozzo2013,aertsgaborasozzo2012,sozzo2013,aertssozzo2013}. %Our 
This `quantum cognition approach' was inspired by %a two decade 
our research on the foundations of quantum theory, the origins of quantum probability and the identification of genuine quantum aspects, such as contextuality, emergence, entanglement, interference, superposition, in macroscopic domains. A `SCoP formalism' was worked out which relies on the interpretation of a concept as an `entity in a specific state changing under the influence of a context' rather than as a `container of instantiations'. This representation of a concept was new with respect to traditional approaches and allowed us to elaborate a quantum representation of the guppy effect explaining at the same time its occurrence in terms of contextual influence. Successively, the mathematical formalism of quantum theory was employed to model the overextension and underextension of membership weights measured by Hampton \cite{h1988a,h1988b}. More specifically, the overextension for conjunctions of concepts measured by Hampton \cite{h1988a} was described as an effect of quantum emergence, interference and superposition, which also play a fundamental role in the description of both overextension and underextension for disjunctions of concepts \cite{h1988b}. Furthermore, a specific conceptual combination experimentally revealed the presence of another genuine quantum effect, namely, entanglement \cite{aertssozzo2011,aertsbroekaertgaborasozzo2013,aertsgaborasozzo2012,aertssozzo2013}. In this paper, we present an elaborate and unified quantum-mechanical representation of concept combinations in Fock space which faithfully agrees with different sets of data collected on concept combinations. Our modeling suggests an explanatory hypothesis according to which human thought is a quantum superposition of an `emergent thought' and a `logical thought', and that the quantum-theoretic approach in Fock space enables this approach to general human thought, consisting of a superposition of these two modes, to be modeled.

\section{Quantum modeling in Fock space\label{brussels}}

Our quantum modeling approach for the combination of two concepts is set in a Fock space $\cal F$ which consists of two sectors: `sector 1' is a Hilbert space $\cal H$, while `sector 2' is a tensor product Hilbert space $\cal H \otimes \cal H$. 

Let us now consider the membership weights of exemplars of concepts and their conjunctions/disjunctions measured by Hampton \cite{h1988a,h1988b}. He identified systematic deviations from classical set (fuzzy set) conjunctions/disjunctions, an effect known as `overextension' or `underextension'. 

Let us start from conjunctions. %Relying on research on the foundations of quantum theory and quantum probability, 
It can be shown that a large part of Hampton's data cannot be modeled in a classical probability space satisfying the axioms of Kolmogorov \cite{aerts2009}. Indeed, the membership weights $\mu_{x}(A), \mu_{x}(B)$ and $\mu_{x}(A\ {\rm and}\ B)$ of an exemplar $x$ for the concepts $A$, $B$ and $`A \ {\rm and} \ B'$ can be represented in a classical probability model if and only if the following two conditions are satisfied (see \cite{aerts2009} for a proof) 
\begin{eqnarray} \label{mindeviation}
\Delta_{x}^{c}=\mu_{x}(A\ {\rm and}\ B)-\min(\mu_{x}(A),\mu_{x}(B)) \le 0 \\ \label{kolmogorovianfactorconjunction}
0 \le k_{x}^{c}=1-\mu_{x}(A)-\mu_{x}(B)+\mu_{x}(A\ {\rm and}\ B)
\end{eqnarray}
%where $\Delta_{x}^{c}$ is the \emph{conjunction rule minimum deviation}, and $k_{x}^{c}$ is the \emph{Kolmogorovian conjunction factor}. 
Let us consider a specific example. Hampton estimated the membership weight of {\it Mint} with respect to the concepts {\it Food}, {\it Plant} and their conjunction {\ it Food and Plant} finding $\mu_{Mint}(Food)=0.87$, $\mu_{Mint}(Plant)=0.81$, $\mu_{Mint}(Food \ {\rm and}\ Plant)=0.9$. Thus, the exemplar \emph{Mint} presents overextension with respect to the conjunction \emph{Food and Plant} of the concepts \emph{Food} and \emph{Plant}. We have in this case $\Delta_{x}^{c}=0.09\not\le0$, hence no classical probability model exists for these data. 

Let us now come to disjunctions. Also in this case, a large part of Hampton's data \cite{h1988b} cannot be modeled in a classical Kolmogorovian probability space, due to the following theorem. 
% Indeed, the 
The membership weights $\mu_{x}(A), \mu_{x}(B)$ and $\mu_{x}(A\ {\rm or}\ B)$ of an exemplar $x$ for the concepts $A$, $B$ and $`A \ {\rm or} \ B'$ can be represented in a classical probability model if and only if the following two conditions are satisfied (see \cite{aerts2009} for a proof) 
\begin{eqnarray} \label{maxdeviation}
\Delta_{x}^{d}=\max(\mu_{x}(A),\mu(_{x}B))-\mu_{x}(A\ {\rm or}\ B)\le 0 \\ \label{kolmogorovianfactordisjunction}
0 \le k_{x}^{d}=\mu_{x}(A)+\mu_{x}(B)-\mu_{x}(A\ {\rm or}\ B)
\end{eqnarray}
%where $\Delta_{x}^{d}$ is the \emph{disjunction maximum rule deviation}, and $k_{x}^{d}$ is the \emph{Kolmogorovian disjunction factor}.
Let us again consider a specific example. Hampton estimated the membership weight of {\it Donkey} with respect to the concepts {\it Pet}, {\it Farmyard Animal} and their disjunction {\it Pet or Farmyard Animal} finding $\mu_{Donkey}(Pet)=0.5$, $\mu_{Donkey}(Farmyard \ Animal)=0.9$, $\mu_{Donkey}(Pet \ {\rm or}\ Farmyard \ Animal)=0.7$. Thus, the exemplar \emph{Donkey} presents underextension with respect to the disjunction \emph{Pet or Farmyard Animal} of the concepts \emph{Pet} and \emph{Farmyard Animal}. We have in this case $\Delta_{x}^{d}=0.2\not\le0$, hence no classical probability model exists for these data.

It can be proved that a quantum probability model in Fock space exists for Hampton's data, as follows \cite{aerts2009,aertsbroekaertgaborasozzo2013,aertsgaborasozzo2012}.

Let us start from the conjunction of two concepts. Let $x$ be an exemplar and let $\mu_{x}(A)$, $\mu_{x}(B)$ and $\mu_{x}(A \ {\rm and} \ B)$, $\mu_{x}(A \ {\rm and} \ B)$ be the membership weights of $x$ with respect to the concepts $A$, $B$ and $`A \ \textrm{and} \ B'$, respectively. Let ${\cal F}={\cal H} \oplus ({\cal H} \otimes {\cal H})$ be the Fock space where we represent the conceptual entities. The concepts $A$, $B$ and $`A \ \textrm{and} \ B'$ are represented by the unit vectors $|A_{c}(x)\rangle$, $|B_{c}(x)\rangle$ and $|(A \ \textrm{and} \ B)_{c}(x)\rangle$, respectively, where
\begin{equation}
|(A \ \textrm{and} \ B)_{c}(x)\rangle=m_{c}(x) e^{i\lambda_{c}(x)}|A_{c}(x)\rangle\otimes|B_{c}(x)\rangle+n_{c}(x)e^{i\nu_{c}(x)}{1\over \sqrt{2}}(|A_{c}(x)\rangle+|B_{c}(x)\rangle)
\end{equation}
The numbers $m_{c}(x)$ and $n_{c}(x)$ are such that $m_{c}(x), n_{c}(x)\ge 0$ and $m_{d}^{2}(x)+n_{c}^{2}(x)=1$. The decision measurement of a subject who estimates the membership of the exemplar $x$ with respect to the concept $`A \  \textrm{and} \ B'$ is represented by the orthogonal projection operator $M_{c}\oplus (M_{c} \otimes M_{c})$ on ${\cal F}$, where $M_{c}$ is an orthogonal projection operator on ${\cal H}$. Hence, the membership weight of $x$ with respect to $`A \  \textrm{and} \ B'$ is given by
\begin{eqnarray} \label{AND}
\mu_{x}(A \ \textrm{and} \ B)=\langle (A \ \textrm{and} \ B)_{c}(x)|M_{c} \oplus (M_{c} \otimes M_{c})|(A \ \textrm{and} \ B)_{c}(x) \rangle \nonumber \\
=m_{c}^2(x)\mu_{x}(A)\mu_{x}(B)+n_{c}^2(x) \Big ({\mu_{x}(A)+\mu_{x}(B) \over 2}+\Re\langle A_{c}(x)|M_{c}|B_{c}(x)\rangle \Big )
\end{eqnarray}
The term $\Re\langle A_{c}(x)|M_{c}|B_{c}(x)\rangle$ is called `interference term' in quantum theory, since it is responsible of the deviations from classicality in the quantum double-slit experiment. In \cite{aerts2009,aertsgaborasozzo2012} we have proved that the model above can be realized in the Fock space $\mathbb{C}^{3} \oplus (\mathbb{C}^{3}\otimes \mathbb{C}^{3})$ with the interference term given by
\begin{equation}
\Re\langle A_{c}(x)|M_{c}|B_{c}(x)\rangle=\sqrt{1-\mu_{x}(A)}\sqrt{1-\mu_{x}(B)}\cos\theta_{c}(x)
\end{equation}
with $\theta_{c}(x)$ being the `interference angle for the conjunction', and $M_{c}=|100 \rangle \langle 100|+|010 \rangle \langle 010|$, where $\{|100\rangle, |010\rangle, |001\rangle\}$ is the canonical basis of ${\mathbb C}^{3}$. For example, in the case of {\it Mint} with respect to {\it Food}, {\it Plant} and {\it Food and Plant}, we have $m_{c}^{2}(x)=0.3$, $n_{c}^{2}(x)=0.7$ and $\theta_{c}(x)=50.21^{\circ}$.  

Let us come again to the disjunction of two concepts. The concepts $A$, $B$ and $`A \ \textrm{or} \ B'$ are represented in the Fock space $\cal F$ by the unit vectors $|A_{d}(x)\rangle$, $|B_{d}(x)\rangle$ and $|(A \ \textrm{or} \ B)_{d}(x)\rangle$, respectively, where
\begin{equation}
|(A \ \textrm{or} \ B)_{c}(x)\rangle=m_{d}(x) e^{i\lambda_{d}(x)}|A_{d}(x)\rangle\otimes|B_{d}(x)\rangle+n_{d}(x)e^{i\nu_{d}(x)}{1\over \sqrt{2}}(|A_{d}(x)\rangle+|B_{d}(x)\rangle)
\end{equation}
The numbers $m_{d}(x)$ and $n_{d}(x)$ are such that $m_{d}(x), n_{d}(x)\ge 0$ and $m_{d}^{2}(x)+n_{d}^{2}(x)=1$. The decision measurement of a subject who estimates the membership of the exemplar $x$ with respect to the concept $`A \  \textrm{or} \ B'$ is represented by the orthogonal projection operator $M_{d}\oplus (M_{d} \otimes \mathbbmss{1}+ \mathbbmss{1} \otimes M_{d}+ M_{d} \otimes M_{d})$ on ${\cal F}$, where $M_{d}$ is an orthogonal projection operator on ${\cal H}$. Hence, the membership weight of $x$ with respect to $`A \  \textrm{or} \ B'$ is given by
\begin{eqnarray} \label{OR}
\mu_{x}(A \ \textrm{and} \ B)=\langle (A \ \textrm{or} \ B)_{c}(x)|M_{d}\oplus (M_{d} \otimes \mathbbmss{1}+ \mathbbmss{1} \otimes M_{d}+ M_{d} \otimes M_{d})|(A \ \textrm{or} \ B)_{c}(x) \rangle \nonumber \\
=m_{d}^2(x) (\mu_{x}(A)+\mu_{x}(B)-\mu_{x}(A)\mu_{x}(B))+n_{c}^2(x) \Big ({\mu_{x}(A)+\mu_{x}(B) \over 2}+\Re\langle A_{d}(x)|M_{d}|B_{d}(x)\rangle \Big )
\end{eqnarray}
The term $\Re\langle A_{d}(x)|M_{d}|B_{d}(x)\rangle$ is the `interference term for the disjunction'. In \cite{aerts2009,aertsgaborasozzo2012} we have proved that the model above can be realized in the Fock space $\mathbb{C}^{3} \oplus (\mathbb{C}^{3}\otimes \mathbb{C}^{3})$ with the interference term given by
\begin{equation}
\Re\langle A_{d}(x)|M_{d}|B_{d}(x)\rangle=\sqrt{1-\mu_{x}(A)}\sqrt{1-\mu_{x}(B)}\cos\theta_{d}(x)
\end{equation}
with $\theta_{d}(x)$ being the `interference angle for the disjunction'. Concerning the {\it Donkey} case, we have $m_{d}^2(x)=0.26$, $n_{d}^2(x)=0.74$ and $\theta_{d}(x)=77.34^{\circ}$.

By comparing Equations (\ref{AND}) and (\ref{OR}), we can see that the interference terms are generally different. Indeed, the representation of the unit vectors $|A_{c}(x)\rangle$, $|A_{d}(x)\rangle$, $|B_{c}(x)\rangle$ and $|B_{d}(x)\rangle$ generally depend on the exemplar $x$, on the membership weights $\mu_{x}(A)$ and $\mu_{x}(B)$, and also on whether $\mu_{x}(A \ \textrm{or} \ B)$ or $\mu_{x}(A \ \textrm{and} \ B)$ is measured, which results in different interferences angles $\theta_{c}(x)$ and $\theta_{d}(x)$.

\section{Conclusions\label{conclusions}}

The probabilistic expressions in the previous section allow the modeling of almost all of Hampton's data \cite{h1988a,h1988b}, describing the deviations from classical logic and probability theory in terms of genuine quantum aspects. Moreover, one of us has recently shown \cite{sozzo2013} that our quantum approach successfully models the data collected by Alxatib and Pelletier \cite{ap2011} on the so-called `borderline contradictions', and it can be further tested to model data coming from future cognitive tests.  One can then inquire into the existence of underlying mechanisms determining these deviations from classicality and, conversely, the effectiveness of a quantum-theoretic modeling. Our explanations is the following. 

Whenever two concepts $A$ and $B$ are combined in human thought to form the conjunction $`A \ {\rm and} \ B'$, or the disjunction $`A \ {\rm or} \ B'$, a new genuine effect comes into play, namely emergence. More specifically, if a subject is asked to estimate whether a given exemplar $x$ belongs to the vague concepts $A$, $B$, $`A \ {\rm and} \ B'$ ($`A \ {\rm or} \ B'$), two mechanisms act simultaneously and in superposition in the subject's thought. A `quantum logical thought', which is a probabilistic version of the classical logical reasoning, where the subject considers two copies of exemplar $x$ and estimates whether the first copy belongs to $A$ and (or) the second copy of $x$ belongs to $B$. But also a `quantum conceptual thought' acts, where the subject estimates whether the exemplar $x$ belongs to the newly emergent concept $`A \ {\rm and} \ B'$ ($`A \ {\rm or} \ B'$). The place where these superposed processes can be suitably structured is the Fock space. Sector 1 of Fock space hosts the latter process, while sector 2 hosts the former, while the weights $m_{c}^2(x)$ ($m_{d}^{2}(x)$) and $n_{c}^2(x)$ ($n_{d}^{2}(x)$) measure the amount of `participation' of sectors 2 and 1 for the conjunction (disjunction), respectively. But, what happens in human thought during a cognitive test is a quantum superposition of both processes. The abundance of over- and under- extension effects is a significant clue that the dominant dynamics in human thought is governed by emergence, and that logical reasoning is only secondary, at variance with old established beliefs.

It is interesting to observe that similar deviations from logic and classical probability theory are observed in other areas of cognitive science, e.g., decision making (`prisoner's dilemma', `disjunction effect', `conjunction fallcy') and behavioral economics (`Allais, Ellsberg, Machina paradoxes'). In the above perspective, our explanation for the appearence of these phenomena is that what has been identified a fallacy, an effect or a deviation, is a consequence of the dominant dynamics in human thought which is emergent in nature, while what has been typically considered as a default to deviate from, namely logical reasoning, is a consequence of a secondary dynamics within human thought, which is quantum logical in nature.

\end{document}